\documentclass[journal]{IEEEtran}
\usepackage{balance}
\usepackage{cite}
\usepackage{bbding}
\usepackage{amsmath,amssymb,amsfonts}
\usepackage{algorithmic}
\usepackage{graphicx}
\usepackage{textcomp}
\usepackage{multirow}
\usepackage{latexsym}
\usepackage[normalem]{ulem}
\useunder{\uline}{\ul}{}
\usepackage[linesnumbered,ruled]{algorithm2e}
\usepackage{setspace}
\begin{document}
\title{HigeNet: A Highly Efficient Modeling for Long Sequence Time Series Prediction in AIOps}
\author{Jiajia Li, Feng Tan, Cheng He, Zikai Wang, Haitao Song, Lingfei Wu,~\IEEEmembership{Member,~IEEE} and Pengwei Hu,~\IEEEmembership{Member,~IEEE}
\thanks{J. Li is with the Department of Computer Science and Engineering, Shanghai Jiao Tong University, Shanghai 200240, China; and Shanghai Artificial Intelligence Research Institute, Shanghai 200240, China (e-mail: lijiajia@sjtu.edu.cn).}
\thanks{F. Tan, Z. Wang and H. Song are with the Shanghai Artificial Intelligence Research Institute, Shanghai 200240 , China (e-amils: tf.uestc@gmail.com, wangzikai@sairi.com.cn and Songhaitao@sairi.com.cn).}
\thanks{C. He is with the Shanghai Dingmao Information Technology Inc., Shanghai, China (e-mail: cheng.he@di-matrix.com).}
\thanks{L. Wu is with the Pinterest, New York 10018, NY, USA (e-mail:lwu@email.wm.edu).}
\thanks{P. Hu is with the Science \& Technology Office, Merck KGaA, Darmstadt 64283, Germany (email: hupengwei@hotmail.com).}
\thanks{*Corresponding authors: Pengwei Hu.}
}
\maketitle

\begin{abstract}
Modern IT system operation demands the integration of system software and hardware metrics. As a result, it generates a massive amount of data, which can be potentially used to make data-driven operational decisions. In the basic form, the decision model needs to monitor a large set of machine data, such as CPU utilization, allocated memory, disk and network latency, and predicts the system metrics to prevent performance degradation. Nevertheless, building an effective prediction model in this scenario is rather challenging as the model has to accurately capture the long-range coupling dependency in the Multivariate Time-Series (MTS). Moreover, this model needs to have low computational complexity and can scale efficiently to the dimension of data available. In this paper, we propose a highly efficient model named HigeNet to predict the long-time sequence time series. We have deployed the HigeNet on production in the D-matrix platform. We also provide offline evaluations on several publicly available datasets as well as one online dataset to demonstrate the model's efficacy. The extensive experiments show that training time, resource usage and accuracy of the model are found to be significantly better than five state-of-the-art competing models.
\end{abstract}
\section{Introduction}
\label{sec:intro}
\begin{figure}[t]
\centerline{\includegraphics[width=\linewidth]{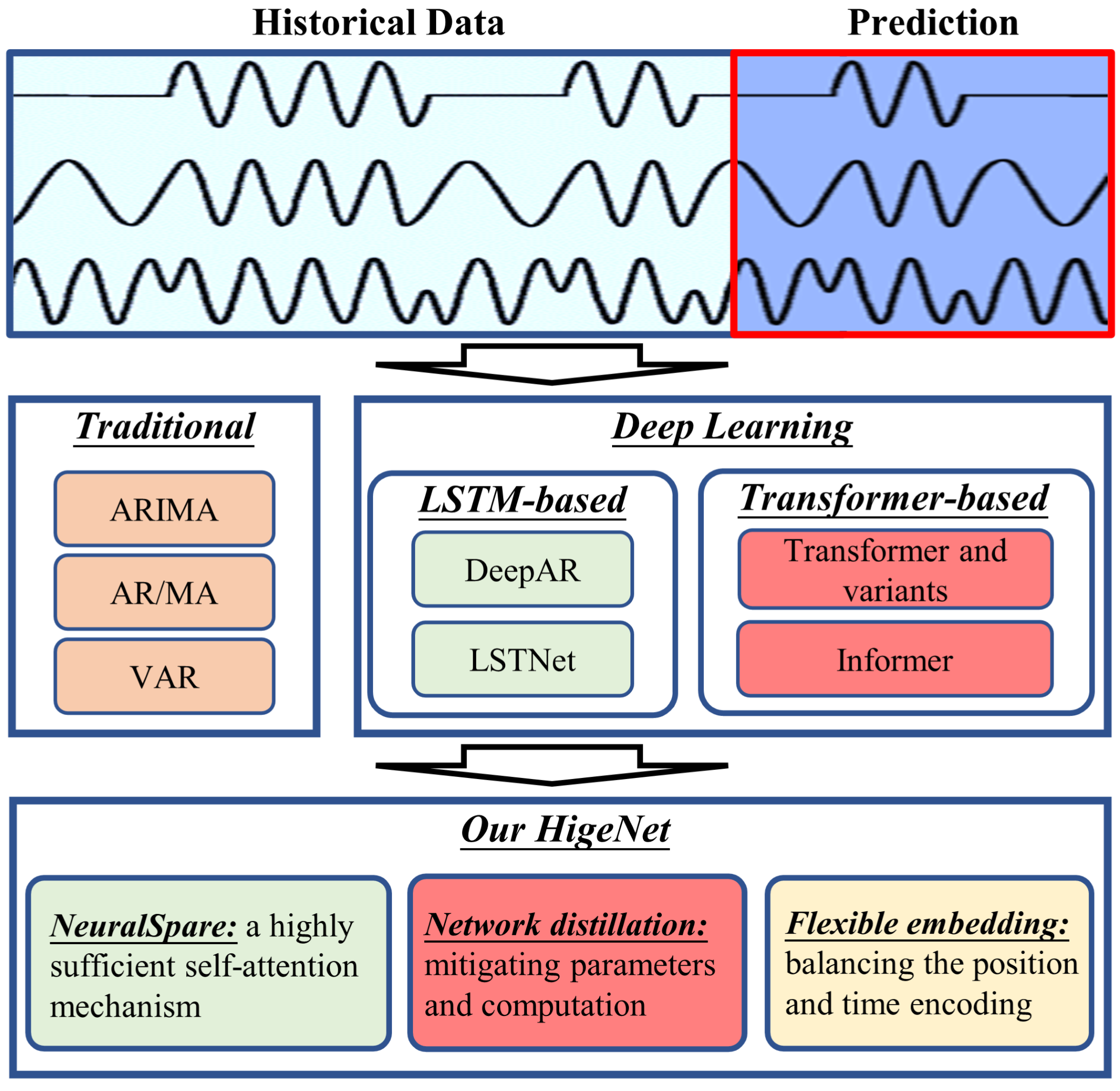}}
\caption{Introduction of existing methods and our HigeNet.}
\label{intro}
\end{figure}
The presence of Information Technology (IT) has transformed the industries and human life. As IT system scales up, its component coupling become more complex, which demands more human resources for its bring up and operation. However, the scale and complexity of the computer systems steadily increase to a level where the manual operation becomes infeasible. As a result, IT operators start employing artificial intelligence tools to deal with the growing operational complexity and costs. 

The main application scenarios of Artificial Intelligence for IT Operations (AIOps) are system bottleneck analysis, anomaly detection, and root cause detection. Datacenter scale system are often service-oriented and contains hundreds of software and hardware components. These components are interconnected and need to respond to the needs of multiple applications. Hence, it generates a large amount of data, which is usually called operation metrics, e.g. CPU/disk/memory and database workload data. Furthermore, the system throughput is a key indicator for the end-to-end system performance and is closely related to the operation of underlying components. Thus, AIOps analysis using these data can effectively help detect, predict and prevent system throughput slowdown that could impact end users \cite{aiops0,aiops2,aiops3}. 

The analysis of the operation data poses several challenges. First, system metrics such as CPU utilization, allocated memory or disk I/O statistics are usually modeled as multivariate time-series (MTS), which requires the ability to capture their long-range coupling dependency. However, the prediction performance of existing methods degrades as sequence length increases.\cite{shortTerm0,shortTerm1,shortTerm2}. Second, monitoring systems have to track the evolution of a large number of time series simultaneously, which often leads to a considerable flow of data to process in near real-time. Therefore, the predictive model needs to have low computational complexity and can scale efficiently to the dimension of data available \cite{KDD1}. Finally, as the time series have different nature, e,g, CPU usage, network latency, database connections, the model has to be flexible such that different characteristics can be accommodated.

In the early days, the problem was solved mainly by manually designed features and statistics (e.g., ARIMA\cite{arim}, AR/MA\cite{arim}, VAR\cite{SVR}). However, as the complexity of data dimension increases, feature engineering cannot be feasible anymore. Thanks to the recent advances in deep learning technology, CNN- and RNN-based \cite{DeepAR, LSTM} deep models are widely used to solve the temporal dependence on sequences, but they cannot be trained in parallel and have disadvantages in long-term dependence. To address this, Transformer\cite{transformer} and its variants become mainstream\cite{ConvTrans,SparseTrans,Reformer,Linformer,Informer}. With the addition of the ST-Norm module, Wavenet\cite{wavenet} and Transformer can achieve excellent forecasting results. On the other hand, large computational space complexity poses challenge to Transformer. The Informer\cite{Informer} improves the time and space complexity of the canonical self-attention\cite{transformer}, however, it still suffers a considerable amount of computation time and resources in computing intermediate variables and multi-layer stacked encoder structures. 

To address the MTS problem in AIOps, this paper propose a highly efficient deep neural network named HigeNet and makes three major contributions:
\begin{itemize}
    \item We propose a novel self-attention mechanism called NeuralSparse, which optimizes computational time and memory complexity by acquiring salient attention dot-product pairs through a learnable neural network.
    \item We design the self-attention distillation mechanism to acquire as many feature features as possible without stacking multiple encoders.
    \item We introduce a new embedding way to fuse and balance the position and time encoding with robust performance. 
\end{itemize}

Additionally, we evaluate our model on a number of data sets including three publicly available and a real-world production data set. The empirical results show that our method compares favorably to classical deep and statistical models. While we determine the best amount of past data needed for further prediction in production settings, we further show that our method is competitive in online prediction task.

\section{Related work}
Statistical approaches such as the Autoregressive Integrated Moving Average (ARIMA) model, autoregressive (AR), Moving Average (MA), and Autoregressive Moving Average (ARMA)\cite{arim,SVR} are the most well-known models that aim at linear univariate time series. However, most of these models are limited to linear univariate time series and do not scale well to MTS that involves multidimensional features. Thus deep-learning-based approaches are emerging, where the convolutional neural networks (CNN) and recurrent neural networks (RNN) are widely applied. For instance, the first proposed DeepAR\cite{DeepAR} aims to produce accurate probabilistic forecasts in time series, based on training an autoregressive recurrent neural network model. In LSTNet\cite{LSTNET}, one-dimensional ordinary convolution is used to capture short-term local information, while GRU \cite{GRU} and Skip-RNN for capturing long-term macroscopic information. In addition, the attention module is added to model the periodicity and the autoregressive process is added in the prediction part. The TPA-LSTM \cite{TPALSTM} improves the attention mechanism by focusing on the selection of key variables. However, these approaches are difficult to be trained in parallel and have disadvantages in long-term dependence. 

Motivated by these challenges, we notice that Transformer-based\cite{transformer} models have been gradually applied to MTS with good results in recent years. As the state-of-the-art model, Transformer\cite{transformer} and its variants are good at capturing long-range dependencies in time series data and solving the gradient disappearance problem effectively. FEDformer\cite{fed} also uses a seasonal-trend decomposition method to capture a global profile with Transformer to capture detailed information. Fourier-enhanced and Wavelet-enhanced blocks are used in FEDformer's Transformer structure so that the attention mechanism can be applied to the frequency domain. Temporal Fusion Transformer(TFT)\cite{tft} was proposed to solve multi-horizon forecasting tasks. It combines gating mechanisms and multi-head self-attention mechanisms. This eliminates unused information and components while enabling memory for long and short-term temporal relationships.
Gated Transformer \cite{gatetn} extends the original Transformer. It builds two towers of Transformer, one step-wise and another channel-wise. Then it merges features of both towers by gating mechanism for MTS classification. Although these transformer-based methods have achieved the comparable precision, however, the large computational space complexity poses challenge to them. Informer\cite{Informer} tried to improve the time and space complexity of the canonical self-attention\cite{transformer}, however, it still suffers a considerable amount of computation time and resources in computing intermediate variables and multi-layer stacked encoder structures. To address the MTS problem, this paper proposes a highly efficient deep neural network named HigeNet evaluated in a scenario of AIOps.

The rest of the paper is organized as follows. In Section 3, we first introduce the problem definition and its background; Section 4 formally describes the model; Section 5 empirically evaluates the model and compares it to the classical deep and statistical models; Section 6 discusses and concludes the paper. 

\section{Problem Definition}
Generally, the analysis of the operation metrics is defined as a MTS prediction problem \cite{KDD1, Informer,TPALSTM,LSTNET,ARIMA,shortTerm0,shortTerm1,shortTerm2,Linformer,Reformer}. The input in MTS can be defined as:
\begin{equation}
    X^t = x^t_1,...,x^t_{L_x}|x_i^t \in \mathbb{R}^{d_x}
\end{equation}, where $t$ means time stamp and $\mathbb{R}^{d_x}$ is the feature dimensions of MTS, the length of MTS is represented as $L_x$. And the output is to predict $Y^{t+h+1} = {y^{t+h+1}_1,...,y^{t+h+1}_{L_y}|y_i^{t+h+1} \in \mathbb{R}^{d_y}}$, where $h$ is the desirable horizon ahead of the current time stamp.

In most cases, parameters are dictated by the specific context, for example, in stock forecasting, the granularity of the forecast needs to be on a second/minute scale. And an hour/day scale is common in weather forecasting. Moreover, when splitting the dataset, the input length $L_x$ and forecast length $L_y$ should be decided. When $d_x$ or $d_y$ is equals to 1, it boils down to the canonical Univariate Time-series problem. In addition, in conventional machine learning or statistical methods, we usually use some dimensionality reduction methods, such as MDS\cite{MDS}, PCA\cite{PCA} and LDA\cite{LDA}, and then carry out the related downstream tasks afterwards. However, in deep learning models, parameters are adaptive and can be learned, which makes the form of model input richer and more sophisticated.

\section{Methodology}
\begin{figure}[t]
\centerline{\includegraphics[width=\linewidth]{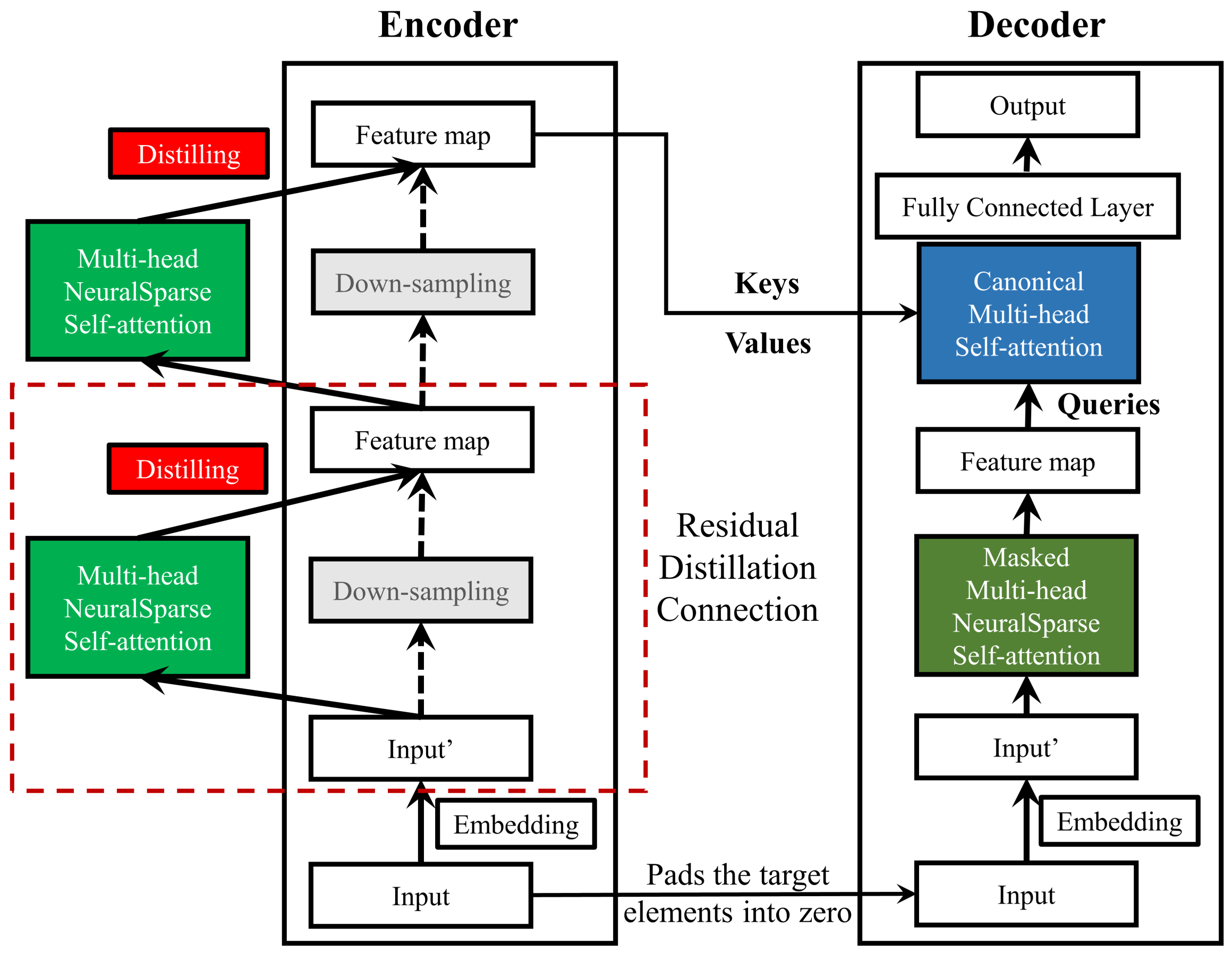}}
\caption{HigeNet model overview. This model uses the popular Encoder-Decoder structure and replaces the predicting way of step-by-step in the Transformer with a generative style. With the architecture as shown in the figure, the model performs up to state-of-the-art.}
\label{arch}
\end{figure}
The overall structure of our HigeNet is shown in Fig. \ref{arch}, which is based on the Encoder-Decoder architecture. 
In the encoding process, the input is embedded and then enters the proposed Multi-head NeuralSparse self-attention module. Compared to conventional self-attention, this module has significant gains in training time and resource consumption, which will be described in details later. The red rectangle indicates the distilling operation that extracts dominating attention and reduces the network size sharply. The dotted box denotes the residual distillation connection. Decoder receives long sequence inputs, pads the target elements into zero and measures the weighted attention composition of the feature map. The pink rectangle represents the feature map generated after the Masked Multi-head NeuralSparse self-attention, which is fed into canonical Multi-head self-attention block as queries with keys and values from encoder.  The fully connected layer instantly predicts output elements in a generative style.

\subsection{Embedding}
Time series is a sequence of data in time order, with values in continuous space. In most cases, raw data is used as the model input after embedding. Therefore, the embedding determines how well the data is represented\cite{Embedding0}. Previous work design the data embedding by manually incorporating different time windows, lag operators, and other manual feature derivations\cite{Embedding1,Embedding2,Embedding3,Embedding4}. However, this approach is too cumbersome and requires domain-specific knowledge. 

In deep learning models, embedding methods based on neural network have been widely used\cite{deepEmbedd0,deepEmbedd1,deepEmbedd2,KDD2,KDD3,KDD4}. In particular, considering the positional semantics and timestamp information can have an impact on the embedding of the data. The 
vanilla transformer\cite{transformer} uses point wise self-attention mechanism and timestamp as local positional context. The Informer\cite{Informer} improves the input representation by adding the global information, e.g. hierarchical timestamps and agnostic timestamps. These approaches establish a way to encode both positional semantics and timestamp. However, the fusion of context features lacks of interpretability\cite{multimodal}. Therefore, the inclusion of position and time encoding is meaningful to the data representation is unclear. 

In this section, we introduce a new embedding way to fuse the position and time encoding that can be reasonably explained. Firstly, we assume that the representation vector at a moment is $[e_1,...,e_d]$, where $d$ is the dimension of the vector. Then, for the first and second occurrences of this representation vector in a time sequence, we set the position vectors $[p_1,...,p_d]$ and $[p_1',...,p_d']$, respectively. Without the position vectors, the angle $\alpha$ between the same two representation vectors at different positions is obviously 0, as shown:
\begin{equation}
\cos (\alpha)=\frac{\sum_{i=1}^{d} e_{i}^{2}}{\sqrt{\left(\sum_{j=1}^{d} e_{j}^{2}\right)\left(\sum_{j=1}^{d} e_{j}^{2}\right)}}=1.
\end{equation}
Furthermore, if the position vectors and the representation vectors are summed up correspondingly, the angle $\alpha$ depends on the difference in position:
\begin{equation}
\cos (\alpha)=\frac{\sum_{i=1}^{d}\left(e_{i}+p_{i}\right)\left(e_{i}+p_{i}^{\prime}\right)}{\sqrt{\left(\sum_{j=1}^{d}\left(e_{j}+p_{j}\right)^{2}\right)\left(\sum_{j=1}^{d}\left(e_{j}+p_{j}\right)^{2}\right)}}.
\end{equation}
\begin{figure*}[t]
\centerline{\includegraphics[width=\linewidth]{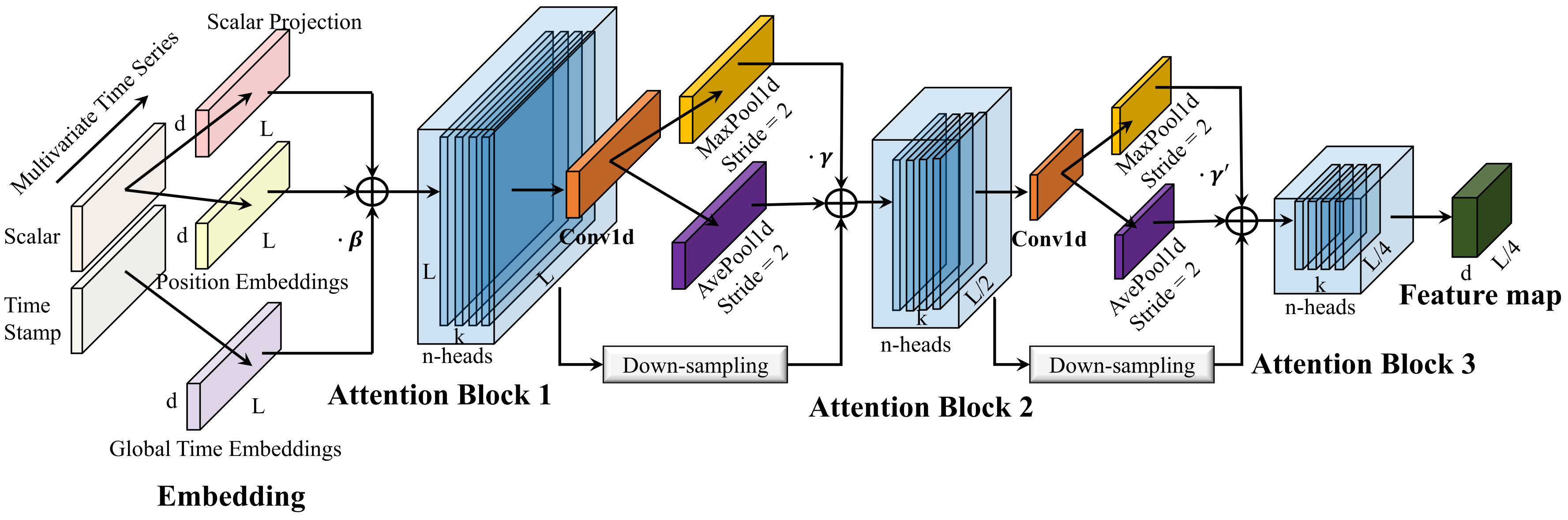}}
\caption{Our HigeNet's encoder. (1) The horizontal stack stands for the encoder in Fig. \ref{arch}. (2) The blue layers are dot-product matrix, and they get cascade decrease by applying self-attention distilling on each layer. (3) $L$ is the length of the input sequence, $d$ is the feature dimension. $\beta$, $\gamma$ and $\gamma$' are the learnable parameters, which is explained in the paper.}
\label{encoder}
\end{figure*}
\begin{figure}[t]
\centerline{\includegraphics[width=\linewidth]{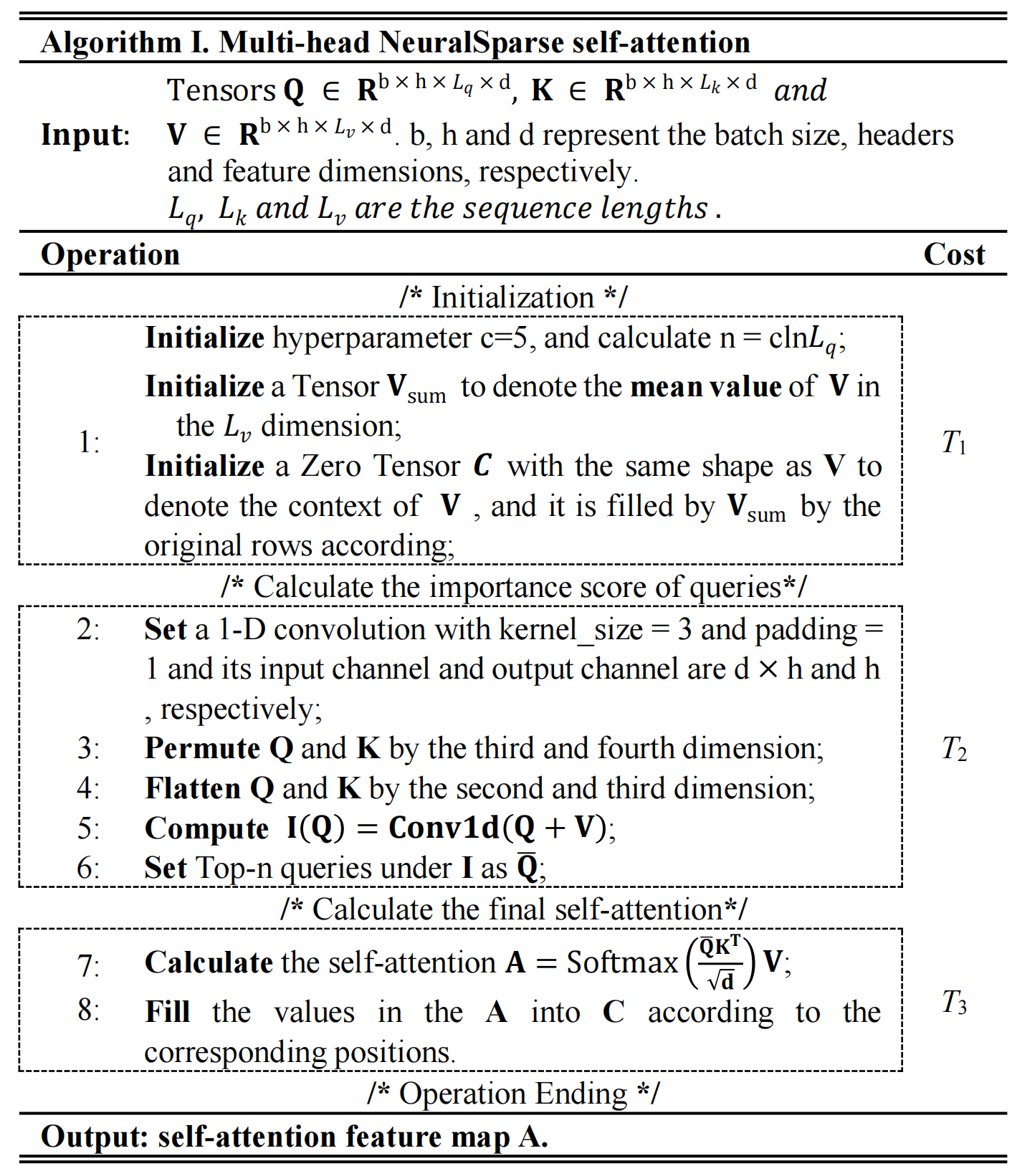}}
\label{NeuralSparse}
\end{figure}

Therefore, this procedure achieves the incorporation of original representation information and positional context. Similarly, to incorporate temporal features, we project the time stamp to the same dimension and embed it directly. Nevertheless, in the actual time series, the importance of local position semantics and global time semantics may differ due to some unexpected events or holidays. To solve this, we design a scaling parameter $\beta$, which is calculated as shown below:
\begin{equation}
    \beta_i^t = Relu(FC(\mathrm{PE}_{\left(L_{x} \times(t-1)+i,\right)}+\sum_{p}\left[\mathrm{SE}_{\left(L_{x} \times(t-1)+i\right)}\right]_{p}))
\end{equation},
where $Relu(\cdot)$ is an activation function, $FC(\cdot)$ denotes the fully connected operation and the input channel number in the $FC$ layer is the feature dimensions ($D_{mdoel}$) and the output channel number is 1.
$\mathrm{PE}_{\left(L_{x} \times(t-1)+i,\right)}$ means the local context, which is embedded as:
\begin{equation}
\begin{aligned}
\operatorname{PE}_{(pos, 2j)} &=\sin \left(pos /\left(2 L_{x}\right)^{2 j / d_{\text {model}}}\right) \\
\operatorname{PE}_{(pos, 2j+1)} &=\cos \left(pos /\left(2 L_{x}\right)^{2 j / d_{\text {model }}}\right)
\end{aligned}
\label{PosEmbedding}
\end{equation},
where $j \in\left\{1, \ldots,\left\lfloor d_{\text {model }} / 2\right\rfloor\right\}$. In other words, once the length $L_x$ of the input sequence and the feature dimension $d_{model}$ have been determined, the position embedding (local context) is fixed. Each global time stamp is employed by a learnable stamp embeddings $SE_{(pos)}$ with limited vocabulary size \cite{Informer}. The final embedding can be expressed as follows:
\begin{equation}
\mathcal{X}_{\text {feed}[i]}^{t}=\mathbf{u}_{i}^{t}+\mathrm{PE}_{\left(L_{x} \times(t-1)+i,\right)}+\beta_i^t\sum_{p}\left[\mathrm{SE}_{\left(L_{x} \times(t-1)+i\right)}\right]_{p}.
\end{equation}
, where $\mathcal{X}_{\text {feed}[i]}^{t}$ is the result of embedding , $i\in\{1,...,L_x\}$ . $\mathbf{u}_{i}^{t}$ denotes the context vector with $d_{model}$-dim after $x_{i}^t$ is projected by 1-D convolutional filters and $d_{model}$ is the feature dimension after input representation. Such an embedding approach not only mines more features of the MTS, but also facilitates training.

\subsection{NeuralSparse Self-attention Mechanism}
The canonical self-attention \cite{transformer} is based on the tuple inputs, i.e, query, key, and value, which can be described as:
$\mathcal{A}(\mathbf{Q}, \mathbf{K}, \mathbf{V})=\operatorname{Softmax}\left(\frac{\mathbf{Q} \mathbf{K}^{T}}{\sqrt{d_{k}}}\right) V $, where $\mathbf{Q}$, $K$ and $\mathbf{V}$ are the matrices of queries, keys and values, respectively. $d_k$ is the input dimension. Further, if $\mathbf{q}_i$, $\mathbf{k}_i$ and $\mathbf{v}_i$ are used to represent the $i$-th row in the $\mathbf{Q}$, $K$ and $\mathbf{V}$ matrix, then the $i$-th row of the output can be expressed as:
\begin{equation}
\mathcal{A}\left(\mathbf{q}_{i}, \mathbf{K}, \mathbf{V}\right)=\sum_{j} \frac{k\left(\mathbf{q}_{i}, \mathbf{k}_{j}\right)}{\sum_{l} k\left(\mathbf{q}_{i}, \mathbf{k}_{l}\right)} \mathbf{v}_{j}=\mathbb{E}_{p\left(\mathbf{k}_{j} \mid \mathbf{q}_{i}\right)}\left[\mathbf{v}_{j}\right]
\end{equation},
where $k\left(\mathbf{q}_{i}, \mathbf{k}_{j}\right)$ is actually an asymmetric exponential kernel function $\exp \left(\frac{q_{i} k_{j}^{T}}{\sqrt{d}}\right)$\cite{expTrans}. It also means that it can weight the summation of the value vectors ($\mathbf{V}$ matrix). It requires the quadratic times dot-product computation and $O(L_QL_K)$ memory usage, which is the major limitation when scaling prediction capacity. 
\begin{table*}[t]
% \small
\renewcommand{\arraystretch}{0.9}
\caption{Univariate long sequence time-series forecasting results on two datasets (four cases). "Count" is the sum of the bolded and underline numbers, which means the optimal and suboptimal performance.}
\label{uni}
\centering
\setlength{\tabcolsep}{0.6mm}
\begin{tabular}{ccccc|ccc|ccc|ccc|ccc}
\hline
\multicolumn{2}{c|}{Method}                                            & \multicolumn{3}{c|}{HigeNet}                     & \multicolumn{3}{c|}{Informer\cite{Informer}}                    & \multicolumn{3}{c|}{TPA-LSTM\cite{TPALSTM}}     & \multicolumn{3}{c|}{LSTNet\cite{LSTNET}} & \multicolumn{3}{c}{ARIMA\cite{ARIMA}} \\ \hline
\multicolumn{2}{c|}{Metric}                                            & CORR           & MSE            & MAE            & CORR           & MSE            & MAE            & CORR        & MSE   & MAE         & CORR    & MSE     & MAE     & CORR    & MSE    & MAE    \\ \hline
\multicolumn{1}{c|}{\multirow{5}{*}{ETTh1}} & \multicolumn{1}{c|}{24}  & {\ul 0.701}    & \textbf{0.073} & \textbf{0.215} & \textbf{0.741} & {\ul 0.098}    & {\ul 0.247}    & 0.598       & 0.123 & 0.331       & 0.635   & 0.123   & 0.331   & 0.558   & 0.123  & 0.331  \\
\multicolumn{1}{c|}{}                       & \multicolumn{1}{c|}{48}  & {\ul 0.414}    & \textbf{0.101} & \textbf{0.255} & \textbf{0.470} & {\ul 0.158}    & {\ul 0.319}    & 0.407       & 0.207 & 0.572       & 0.384   & 0.207   & 0.572   & 0.332   & 0.207  & 0.572  \\
\multicolumn{1}{c|}{}                       & \multicolumn{1}{c|}{168} & \textbf{0.213} & \textbf{0.177} & {\ul 0.349}    & {\ul 0.193}    & {\ul 0.183}    & \textbf{0.346} & 0.187       & 0.458 & 0.629       & 0.205   & 0.458   & 0.629   & 0.103   & 0.458  & 0.629  \\
\multicolumn{1}{c|}{}                       & \multicolumn{1}{c|}{336} & \textbf{0.269} & \textbf{0.183} & \textbf{0.350} & 0.196          & {\ul 0.222}    & {\ul 0.387}    & {\ul 0.239} & 0.505 & 0.759       & 0.157   & 0.505   & 0.759   & 0.077   & 0.505  & 0.759  \\
\multicolumn{1}{c|}{}                       & \multicolumn{1}{c|}{720} & \textbf{0.551} & \textbf{0.123} & \textbf{0.282} & 0.305          & {\ul 0.269}    & {\ul 0.435}    & {\ul 0.387} & 0.733 & 0.804       & 0.222   & 0.733   & 0.804   & 0.039   & 0.733  & 0.804  \\ \hline
\multicolumn{1}{c|}{\multirow{5}{*}{ETTh2}} & \multicolumn{1}{c|}{24}  & \textbf{0.878} & \textbf{0.106} & {\ul 0.253}    & {\ul 0.857}    & {\ul 0.093}    & \textbf{0.240} & 0.802       & 0.195 & 0.355       & 0.823   & 0.133   & 0.290   & 0.528   & 2.377  & 0.345  \\
\multicolumn{1}{c|}{}                       & \multicolumn{1}{c|}{48}  & \textbf{0.815} & \textbf{0.149} & \textbf{0.304} & {\ul 0.796}    & {\ul 0.155}    & {\ul 0.314}    & 0.722       & 0.198 & 0.374       & 0.733   & 0.188   & 0.347   & 0.324   & 3.153  & 0.484  \\
\multicolumn{1}{c|}{}                       & \multicolumn{1}{c|}{168} & \textbf{0.631} & {\ul 0.240}    & {\ul 0.399}    & {\ul 0.617}    & \textbf{0.232} & \textbf{0.389} & 0.487       & 0.338 & 0.463       & 0.544   & 0.316   & 0.450   & 0.273   & 2.987  & 0.567  \\
\multicolumn{1}{c|}{}                       & \multicolumn{1}{c|}{336} & \textbf{0.581} & \textbf{0.258} & \textbf{0.415} & {\ul 0.579}    & {\ul 0.263}    & {\ul 0.417}    & 0.448       & 0.322 & 0.457       & 0.512   & 0.337   & 0.468   & 0.225   & 3.573  & 0.721  \\
\multicolumn{1}{c|}{}                       & \multicolumn{1}{c|}{720} & \textbf{0.561} & \textbf{0.246} & \textbf{0.406} & {\ul 0.534}    & {\ul 0.277}    & {\ul 0.431}    & 0.340       & 0.307 & 0.448       & 0.469   & 0.264   & 0.417   & 0.102   & 3.793  & 0.934  \\ \hline
\multicolumn{1}{c|}{\multirow{5}{*}{ETTm1}} & \multicolumn{1}{c|}{24}  & \textbf{0.947} & \textbf{0.023} & \textbf{0.117} & {\ul 0.945}    & {\ul 0.030}    & {\ul 0.137}    & 0.812       & 0.063 & 0.166       & 0.929   & 0.029   & 0.133   & 0.809   & 0.083  & 0.148  \\
\multicolumn{1}{c|}{}                       & \multicolumn{1}{c|}{48}  & {\ul 0.897}    & \textbf{0.032} & \textbf{0.138} & \textbf{0.902} & {\ul 0.069}    & {\ul 0.203}    & 0.793       & 0.155 & 0.298       & 0.860   & 0.047   & 0.168   & 0.783   & 0.164  & 0.311  \\
\multicolumn{1}{c|}{}                       & \multicolumn{1}{c|}{96}  & \textbf{0.744} & \textbf{0.168} & \textbf{0.350} & {\ul 0.739}    & {\ul 0.194}    & {\ul 0.372}    & 0.633       & 0.229 & 0.392       & 0.705   & 0.184   & 0.476   & 0.623   & 0.249  & 0.427  \\
\multicolumn{1}{c|}{}                       & \multicolumn{1}{c|}{288} & \textbf{0.453} & \textbf{0.187} & \textbf{0.363} & {\ul 0.410}    & {\ul 0.401}    & 0.554          & 0.337       & 0.405 & {\ul 0.538} & 0.354   & 0.294   & 0.562   & 0.328   & 0.407  & 0.593  \\
\multicolumn{1}{c|}{}                       & \multicolumn{1}{c|}{672} & \textbf{0.451} & \textbf{0.421} & \textbf{0.582} & {\ul 0.314}    & {\ul 0.512}    & {\ul 0.644}    & 0.184       & 0.623 & 0.836       & 0.209   & 0.422   & 0.586   & 0.211   & 0.583  & 0.792  \\ \hline
\multicolumn{1}{c|}{\multirow{3}{*}{AIOps}} & \multicolumn{1}{c|}{96}  & {\ul 0.266}    & \textbf{1.296} & {\ul 0.324}    & \textbf{0.270} & {\ul 1.315}    & \textbf{0.306} & 0.173       & 2.384 & 1.635       & 0.227   & 1.358   & 0.658   & 0.198   & 2.073  & 1.173  \\
\multicolumn{1}{c|}{}                       & \multicolumn{1}{c|}{288} & \textbf{0.287} & \textbf{1.300} & \textbf{0.282} & {\ul 0.270}    & {\ul 1.328}    & {\ul 0.283}    & 0.127       & 2.974 & 2.073       & 0.158   & 1.546   & 0.754   & 0.105   & 2.537  & 1.937  \\
\multicolumn{1}{c|}{}                       & \multicolumn{1}{c|}{576} & \textbf{0.277} & \textbf{1.328} & \textbf{0.298} & {\ul 0.253}    & {\ul 1.350}    & {\ul 0.301}    & 0.072       & 3.625 & 2.474       & 0.139   & 1.673   & 0.953   & 0.099   & 3.173  & 2.647  \\ \hline
\multicolumn{2}{c|}{Count(bolded)}                                     & \multicolumn{3}{c|}{45}                          & \multicolumn{3}{c|}{9}                           & \multicolumn{3}{c|}{0}            & \multicolumn{3}{c|}{0}      & \multicolumn{3}{c}{0}     \\ \hline
\multicolumn{2}{c}{Count(underline)}                                   & \multicolumn{3}{c|}{9}                           & \multicolumn{3}{c|}{42}                          & \multicolumn{3}{c|}{3}            & \multicolumn{3}{c|}{0}      & \multicolumn{3}{c}{0}     \\ \hline
\end{tabular}
\end{table*}
\begin{table}[t]
\caption{Excluding the data column, the overall properties of the dataset, where L is the length of the time series, D is the dimension number
of time series, I is the sampling spacing, S is size of the dataset in bytes and M and V are the dataset's mean and variance, respectively. }
\label{our_dataset}
\centering
\setlength{\tabcolsep}{1.0mm}
\begin{tabular}{ccccc}
\hline
\textbf{Features} & \textbf{Whole} & \textbf{Training} & \textbf{Validation} & \textbf{Testing} \\ \hline
\textbf{L}        & 101583                 & 60949                     & 20317                       & 20317                    \\
\textbf{D}        & 20                     & 20                        & 20                          & 20                       \\
\textbf{I}        & 5 mins                 & 5 mins                    & 5 mins                      & 5 mins                   \\
\textbf{S}        & 13309 KB               & 7957 KB                   & 2665 KB                     & 2687 KB                  \\
\textbf{M}     & 857.2778               & 804.7236                  & 866.8692                    & 1005.3438                \\
\textbf{V}      & 2678.4322              & 2498.7410                 & 2670.6132                   & 3158.5293                \\ \hline
\end{tabular}
\end{table}
Numerous studies have shown that sparse self-attention scores form a long-tail distribution, i.e., a few dot product pairs contribute to the main attention and other dot product pairs can be ignored. In this case, if we can calculate the most significant Top-n queries by the relationship between $\mathbf{Q}$ and $\mathbf{K}$, the complexity of $O(L_QL_K)$ can be optimized.
Therefore, we propose a way to implement the filtering process of queries through neural network learning, defined as follows:
\begin{equation}
\mathcal{I}\left(\mathbf{Q}\right)=
Conv1d(\mathbf{Q} + \mathbf{K})
\end{equation},
where $\mathcal{I}\left(\mathbf{Q}\right)$ means the importance score of queries, which has a shape of $L_Q$ $\times$ 1. 
$Conv1d(\cdot)$ is a 1-D convolution (kernel\_size=3, padding=1), in which the feature dimensions of $\mathbf{Q}$ are used as the number of input channels and we use the number of headers from the attention block as the output channels. 
In above way, we discard the method of calculating attention scores in Transformer\cite{transformer}. Instead, we project the features after $\mathbf{Q}$ and $\mathbf{K}$ fusion into a new probability space by using a convolutional layer. Moreover, we obtain the scores of queries and implement the process by selecting the top-$n$ queries with the highest scores by setting $n$=$c$ln$L_Q$. We name this method NeuralSparse self-attention, which is shown in Algorithm \ref{NeuralSparse}. The overall time cost of Algorithm I comes to: $T = T_1 + T_2 + T_3$. In doing this, time and space complexity in self-attention module can be improved from $O(L^2)$ to $O(LlnL)$. 
Noteworthy, Informer also achieves the $O(LlnL)$ complexity, in which the authors concluded that the salient dot product pairs lead the attention probability distribution of the corresponding query away from the uniform distribution, so Kullback-Leibler divergence\cite{KL} was used to evaluate both distributions. Although this approach reduces the complexity of self-attention, it requires a large number of intermediate variables and computational steps in computing the distribution, which has disadvantages in terms of training time and computational resources. 

To summarize, our method has the following advantages: 1. Reduces the computational effort in the process of filtering queries. 2. Obtains faster training speed and lower GPU usage. 3. Achieves good continuity in feature domain.

\subsection{Encoder and Decoder}
\begin{table*}[t]
% \small
\renewcommand{\arraystretch}{0.9}
\caption{Multivariate long sequence time-series forecasting results on two datasets (four cases).}
\label{multi}
\centering
\setlength{\tabcolsep}{0.6mm}
\begin{tabular}{cc|lll|lll|ccc|ccc|ccc}
\hline
\multicolumn{2}{c|}{Method}                       & \multicolumn{3}{c|}{HigeNet}                                                                                  & \multicolumn{3}{c|}{Informer\cite{Informer}}                                                                        & \multicolumn{3}{c|}{Wavenet+STN\cite{wavenet}}                 & \multicolumn{3}{c|}{TPA-LSTM\cite{TPALSTM}} & \multicolumn{3}{c}{LSTNet\cite{LSTNET}} \\ \hline
\multicolumn{2}{c|}{Metric}                       & \multicolumn{1}{c}{CORR}           & \multicolumn{1}{c}{MSE}            & \multicolumn{1}{c|}{MAE}            & \multicolumn{1}{c}{CORR}        & \multicolumn{1}{c}{MSE}         & \multicolumn{1}{c|}{MAE}         & CORR           & MSE            & MAE            & CORR     & MSE      & MAE     & CORR    & MSE     & MAE    \\ \hline
\multicolumn{1}{c|}{\multirow{5}{*}{ETTh1}} & 24  & \textbf{0.642}                     & \textbf{0.487}                     & \textbf{0.504}                      & {\ul 0.628}                     & {\ul 0.577}                     & {\ul 0.549}                      & 0.443          & 0.789          & 0.637          & 0.509    & 1.263    & 0.903   & 0.579   & 1.152   & 0.873  \\
\multicolumn{1}{c|}{}                       & 48  & {\ul 0.559}                        & \textbf{0.627}                     & \textbf{0.588}                      & \textbf{0.566}                  & {\ul 0.685}                     & {\ul 0.625}                      & 0.418          & 0.796          & 0.645          & 0.426    & 1.426    & 1.027   & 0.501   & 1.376   & 0.893  \\
\multicolumn{1}{c|}{}                       & 168 & \textbf{0.216}                     & 1.111                              & 0.851                               & 0.108                           & \textbf{0.931}                  & \textbf{0.752}                   & {\ul 0.141}    & {\ul 1.069}    & {\ul 0.772}    & 0.063    & 2.375    & 1.832   & 0.090   & 1.876   & 1.127  \\
\multicolumn{1}{c|}{}                       & 336 & \textbf{0.220}                     & {\ul 1.241}                        & {\ul 0.903}                         & {\ul 0.150}                     & \textbf{1.128}                  & \textbf{0.873}                   & 0.135          & 1.343          & 1.001          & 0.104    & 2.648    & 2.384   & 0.127   & 2.478   & 1.235  \\
\multicolumn{1}{c|}{}                       & 720 & {\ul 0.139}                        & {\ul 1.390}                        & {\ul 0.959}                         & \textbf{0.148}                  & \textbf{1.215}                  & \textbf{0.896}                   & 0.101          & 1.437          & 1.084          & 0.004    & 4.3OI    & 3.355   & 0.065   & 2.565   & 1.353  \\ \hline
\multicolumn{1}{c|}{\multirow{5}{*}{ETTh2}} & 24  & \textbf{0.562}                     & \textbf{0.355}                     & \textbf{0.462}                      & {\ul 0.548}                     & 0.720                           & 0.665                            & 0.505          & {\ul 0.567}    & {\ul 0.566}    & 0.203    & 2.694    & 1.684   & 0.278   & 2.475   & 1.476  \\
\multicolumn{1}{c|}{}                       & 48  & \textbf{0.107}                     & \textbf{1.142}                     & \textbf{0.998}                      & {\ul 0.088}                     & {\ul 1.457}                     & {\ul 1.001}                      & 0.074          & 2.443          & 1.243          & 0.014    & 3.656    & 1.583   & 0.007   & 3.874   & 1.639  \\
\multicolumn{1}{c|}{}                       & 168 & 0.083                              & {\ul 4.086}                        & {\ul 1.742}                         & \textbf{0.140}                  & \textbf{3.489}                  & \textbf{1.515}                   & {\ul 0.085}    & 4.387          & 1.836          & 0.047    & 4.876    & 2.984   & 0.118   & 4.139   & 2.683  \\
\multicolumn{1}{c|}{}                       & 336 & {\ul 0.061}                        & 3.062                              & 1.497                               & 0.055                           & {\ul 2.723}                     & {\ul 1.340}                      & \textbf{0.156} & \textbf{2.228} & \textbf{1.235} & 0.042    & 3.384    & 2.837   & 0.060   & 3.099   & 2.715  \\
\multicolumn{1}{c|}{}                       & 720 & \textbf{0.127}                     & \textbf{3.332}                     & \textbf{1.466}                      & {\ul 0.097}                     & {\ul 3.467}                     & {\ul 1.473}                      & 0.011          & 5.385          & 2.475          & 0.023    & 4.983    & 3.832   & 0.022   & 4.786   & 3.666  \\ \hline
\multicolumn{1}{c|}{\multirow{5}{*}{ETTm1}} & 24  & \textbf{0.772}                     & \textbf{0.308}                     & \textbf{0.379}                      & 0.737                           & {\ul 0.323}                     & {\ul 0.369}                      & {\ul 0.759}    & 0.417          & 0.445          & 0.337    & 1.687    & 1.287   & 0.203   & 2.038   & 1.472  \\
\multicolumn{1}{c|}{}                       & 48  & \textbf{0.701}                     & \textbf{0.434}                     & \textbf{0.433}                      & {\ul 0.681}                     & {\ul 0.494}                     & 0.503                            & 0.658          & 0.497          & {\ul 0.484}    & 0.208    & 2.076    & 1.486   & 0.198   & 2.108   & 1.523  \\
\multicolumn{1}{c|}{}                       & 96  & \textbf{0.657}                     & \textbf{0.521}                     & \textbf{0.523}                      & 0.605                           & 0.678                           & 0.614                            & {\ul 0.632}    & {\ul 0.537}    & {\ul 0.529}    & 0.269    & 1.926    & 1.378   & 0.187   & 2.876   & 1.738  \\
\multicolumn{1}{c|}{}                       & 288 & \textbf{0.470}                     & \textbf{0.797}                     & \textbf{0.693}                      & 0.439                           & 1.056                           & 0.786                            & {\ul 0.451}    & {\ul 0.827}    & {\ul 0.705}    & 0.187    & 2.763    & 2.011   & 0.208   & 2.543   & 2.087  \\
\multicolumn{1}{c|}{}                       & 672 & \textbf{0.404}                     & \textbf{0.877}                     & \textbf{0.739}                      & {\ul 0.381}                     & 1.192                           & 0.926                            & 0.373          & {\ul 0.921}    & {\ul 0.787}    & 0.201    & 1.863    & 1.825   & 0.098   & 2.376   & 2.893  \\ \hline
\multicolumn{1}{c|}{\multirow{3}{*}{AIOps}} & 96  & \multicolumn{1}{c}{\textbf{0.606}} & \multicolumn{1}{c}{\textbf{0.624}} & \multicolumn{1}{c|}{\textbf{0.383}} & \multicolumn{1}{c}{{\ul 0.576}} & \multicolumn{1}{c}{{\ul 0.679}} & \multicolumn{1}{c|}{{\ul 0.418}} & 0.518          & 0.727          & 0.454          & 0.372    & 1.273    & 0.876   & 0.505   & 0.756   & 0.574  \\
\multicolumn{1}{c|}{}                       & 288 & \multicolumn{1}{c}{\textbf{0.588}} & \multicolumn{1}{c}{\textbf{0.653}} & \multicolumn{1}{c|}{\textbf{0.373}} & \multicolumn{1}{c}{{\ul 0.556}} & \multicolumn{1}{c}{0.711}       & \multicolumn{1}{c|}{{\ul 0.397}} & 0.540          & {\ul 0.681}    & 0.418          & 0.172    & 2.385    & 1.736   & 0.487   & 0.728   & 0.598  \\
\multicolumn{1}{c|}{}                       & 576 & \multicolumn{1}{c}{\textbf{0.555}} & \multicolumn{1}{c}{\textbf{0.692}} & \multicolumn{1}{c|}{\textbf{0.410}} & \multicolumn{1}{c}{0.525}       & \multicolumn{1}{c}{0.753}       & \multicolumn{1}{c|}{0.461}       & {\ul 0.533}    & {\ul 0.698}    & {\ul 0.437}    & 0.064    & 4.868    & 4.872   & 0.479   & 0.898   & 0.791  \\ \hline
\multicolumn{2}{c|}{Count(bolded)}                & \multicolumn{3}{c|}{40}                                                                                       & \multicolumn{3}{c|}{11}                                                                              & \multicolumn{3}{c|}{3}                           & \multicolumn{3}{c|}{0}        & \multicolumn{3}{c}{0}      \\ \hline
\multicolumn{2}{c|}{Count(underline)}             & \multicolumn{3}{c|}{9}                                                                                        & \multicolumn{3}{c|}{25}                                                                              & \multicolumn{3}{c|}{20}                          & \multicolumn{3}{c|}{0}        & \multicolumn{3}{c}{0}      \\ \hline
\end{tabular}
\end{table*}
\subsubsection{Encoder}
To extract the robust long-range dependency of the long sequential inputs, we propose a single-encoder approach for feature extraction and improve the distillation operation, which is represented in Fig. \ref{encoder}. After the input embedding is calculated by our NeuralSparse self-attention module, we obtain the $n$-heads weights matrix (overlapping blue squares) of Attention blocks in Fig. \ref{encoder}. And our "distilling" procedure from $j$-th attention block to ($j$+1)-th attention block can be defined as:
\begin{equation}
\mathbf{X}_{j+1}^{t}=\operatorname{MP}(\left[\mathbf{F}_{j}^{t}\right]_{\mathrm{AB}}) + \mathbf{\gamma\cdot}\operatorname{AP}(\left[\mathbf{F}_{j}^{t}\right]_{\mathrm{AB}}) + \operatorname{DS}([\mathbf{X}_{j}^{t}]_{\mathrm{AB}})
\end{equation},
where $[\cdot]_{AB}$ represents the attention block, $AP(\cdot)$ and $MP(\cdot)$ are the global average pooling and global maximum pooling, $\gamma$ is a learnable parameter. $DS(\cdot)$ means the down-sampling operation and we use the global average pooling (stride = 2) for $DS(\cdot)$. Additionally, $\mathbf{F}_{j}^{t}$ is calculated by:
\begin{equation}
\mathbf{F}_{j}^{t}=\operatorname{ELU}\left(\operatorname{Conv}1d\left(\left[\mathbf{X}_{j}^{t}\right]_{\mathrm{AB}}\right)\right)
\end{equation},
where $(\operatorname{Conv}1d(\cdot)$ performs an 1-D convolutional filters (kernel width = 3) on time dimension with the $ELU(\cdot)$ activation function\cite{transformer}. Although downsampling can reduce the dimensionality of features, some semantic information will be lost. To mitigate this effect, we obtain as much semantic information as possible by adding a max-pooling layer and an average-pooling layer in parallel (stride=2). We also add a learnable $\gamma$ to adjust the importance of these two pooling operations. In addition, to prevent the disappearance of gradients and features, we add a residual connection, as shown in Fig. \ref{arch}. After encoder, the length of the feature map becomes a quarter of the original length. Compared to those approaches by stacking encoders, we have less number of parameters, faster computation, and also get as many features as possible.

\subsubsection{Decoder}
The input of decoder contains two parts, one is the output (Keys and Values) of encoder and the other is the embedding after padding the target elements into zero. Compared to Multi-head NeuralSparse self-attention, Masked Multi-head NeuralSparse masks the future part before calculating the Softmax($\cdot$). Instead of filling lazy queries with $mean(\mathbf{V})$, we fill it with Cumsum, which is the accumulation of $\mathbf{V}$ vectors at all time points before each queries. This filling method can prevent the model from focusing on future information. Finally, the Queries, Keys and Values are passed into the Canonical Multi-head Self-attention to obtain the prediction results. Additionally, the MSE loss is propagated back from the decoder’s outputs across the entire model.
\section{Experiment}
\subsection{Datasets}

$\mathbf{AIOps \ dataset}\footnote{ https://github.com/Torchlight-ljj/AIOPSdataset.}$: To validate our HigeNet, we use the monitoring data of an online AIOps system as the dataset. This dataset has a total of 20 dimensions, as shown in Table. \ref{our_dataset}. We divide the dataset into training set, validation set and testing set at the ratio of 6:2:2.

$\mathbf{ETT \ dataset}\footnote{https://github.com/zhouhaoyi/ETDataset.}$(Electricity Transformer Temperature): To demonstrate the generalization performance of our model, we have selected another publicly available dataset, which is separated into three sub-datasets according to the different time granularity. Among these sub-datasets, ETTh$_1$ and ETTh$_2$ are for 1-hour-level and ETTm$_1$ is for 15-minute-level. We divide the datasets into training set, validation set and testing set at the ratio of 6:2:2.

\subsection{Experimental Settings}
\begin{figure*}[t]
\centerline{\includegraphics[width=\linewidth]{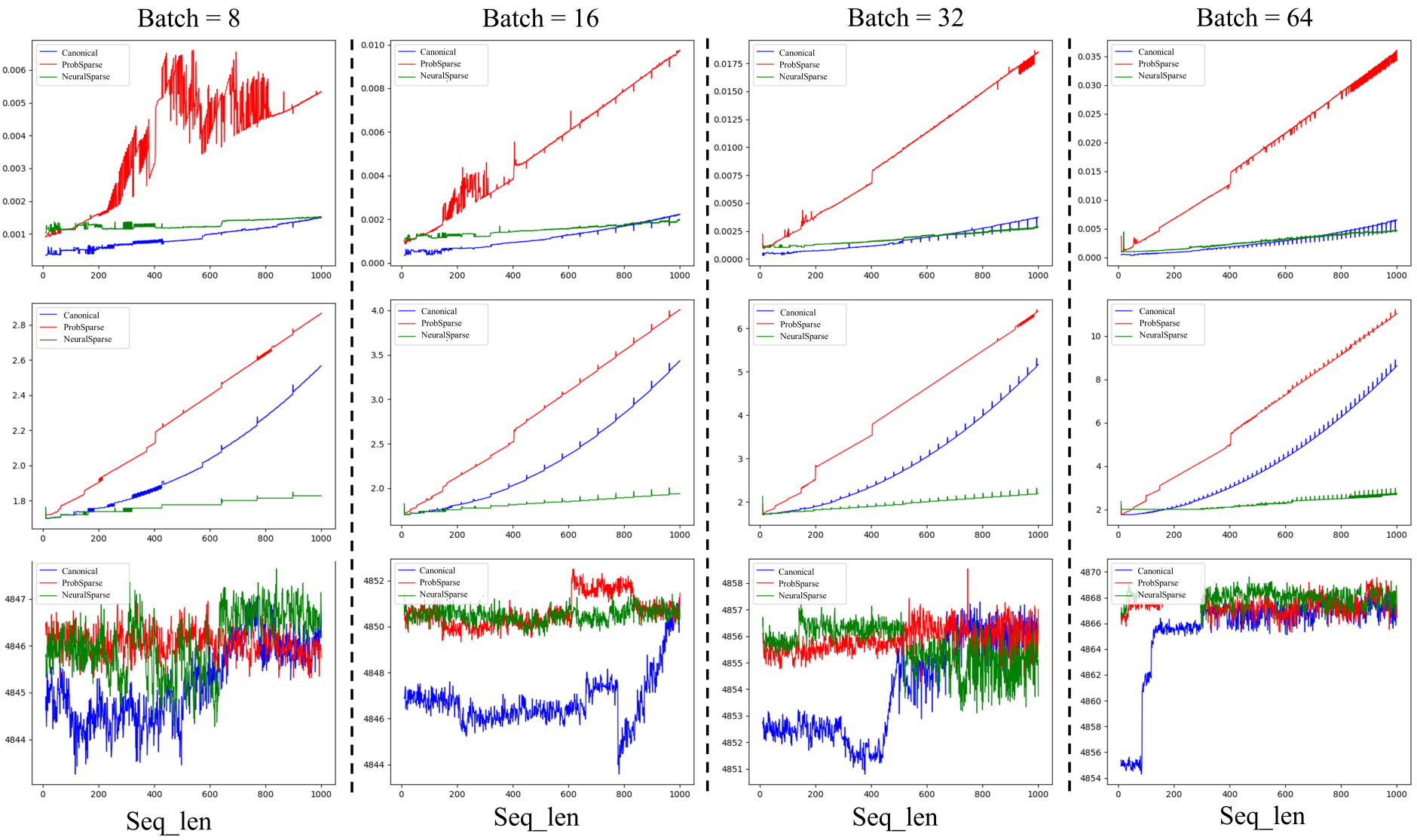}}
\caption{The performance comparison of different self-attention modules. The horizontal coordinate of each of these sub-images is the length of the input sequence. The first, second and third rows indicate the computation time (s), GPU resource usage (GB) and memory usage (MB), respectively. The green, red and blue lines represent the testing results of NeuralSparse, ProbSparse from Informer and Canonical from Transformer self-attention modules, respectively. }
\label{self-attentionComps}
\end{figure*}

All methods are implemented in the Pytorch framework $\footnote{https://pytorch.org/}$. Our HigeNet use Adam optimizer for training with initial learning rate $10^{-4}$, weight decay 5$e^{-4}$, momentum 0.9, batch size 32, and iteration 20 epochs. The learning rate is decayed by 0.5 every 5 epochs. The training is implemented on one NVIDIA Geforce GTX 3090 Ti GPU and Intel(R) Core(TM) i9-10900K CPU. The source code is available at https://github.com/Torchlight-ljj/HigeNet.

For the evaluation metrics of the model, we use $CORR$, $MAE$ and $MSE$, where $CORR$ means the empirical correlation coefficient, $MAE$ is Mean Absolute Error and $MSE$ represents Mean Squared Error. They are defined as below:
\begin{equation}
MAE=\frac{1}{n} \sum_{i=1}^{n}\left|\hat{y}_{i}-y_{i}\right|
\end{equation}
\begin{equation}
MSE=\frac{1}{n} \sum_{i=1}^{n}\left(y_{i}-\hat{y}_{i}\right)^{2}
\end{equation}
\begin{equation}
\operatorname{CORR}= \frac{\sum_{i=1}^{n}\left(y_{i}-\operatorname{mean}\left(y\right)\right)\left(\hat{y}_{i}-\operatorname{mean}\left(\hat{y}\right)\right)}{\sqrt{\sum_{i=1}^{n}\left(y_{i}-\operatorname{mean}\left(y\right)\right)^{2}}\sqrt{\sum_{i=1}^{n}\left(\hat{y}_{i}-\operatorname{mean}\left(\hat{y}\right)\right)^{2}}}
\end{equation}
where $y$ and $\hat{y}$ are ground-truth signals and system prediction signals, respectively. Furthermore, we set $y=\{y_1,y_2,...,y_n\}$ and $\hat{y}=\{\hat{y}_1,\hat{y}_2,...,\hat{y}_n\}$ and $n$ means the number of the samples.
We compare HigeNet with the four most dominant current deep models, including TPA-LSTM\cite{TPALSTM}, LSTNet\cite{LSTNET}, Informer\cite{Informer} and Wavenet+STN\cite{KDD3},
and one conventional statistical model ARIMA\cite{ARIMA}.
\subsection{Results and Analysis}

Univariate and multivariate time-series forecasting results are in Table \ref{uni} and Table \ref{multi}. We perform related experiments on two datasets (four cases) and choose different prediction steps while counting the number of optimal and suboptimal performances of each algorithm.
\subsubsection{Univariate Time-series Forecasting}
On the ETT dataset, we use the column "oil temperature" as experimental data. On the AIOps dataset, we set "Response time" as the target value. From Table \ref{uni}, we can observe that: (1) In all four cases, HigeNet outperforms the other four models (count=45>9>0), maintaining a higher $CORR$ and lower $MSE$ and $MAE$. (2) Informer obtains the largest $CORR$s (0.741 and 0.470) at the prediction step 24 and 48 in ETTh$_1$. And its $MAE$ is the lowest when the prediction step size is 168. Moreover, in ETTm$_1$ and AIOps datasets, Informer also has several metrics that stand out the most. (3) For the other three methods, LSTNet outperforms TPA-LSTM and ARIMA achieves the worst results.
\subsubsection{Multivariate Time-series Forecasting}
\begin{figure*}[t]
\centerline{\includegraphics[width=\linewidth]{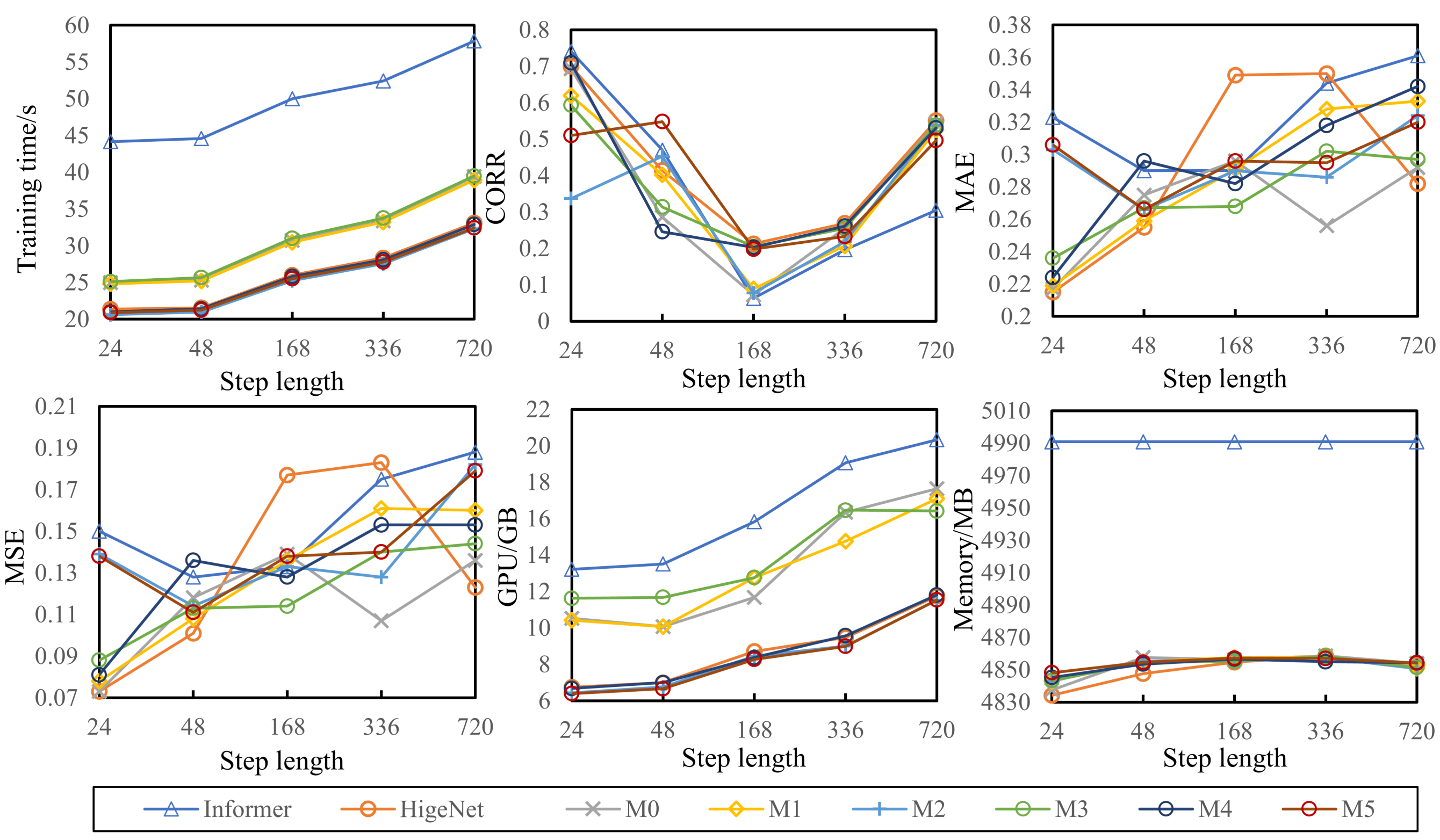}}
\caption{Univariate long sequence time-series forecasting results for the ablation on ETTh$_1$ dataset}
\label{ETTh1_uniablations}
\end{figure*}
\begin{figure*}[t]
\centerline{\includegraphics[width=\linewidth]{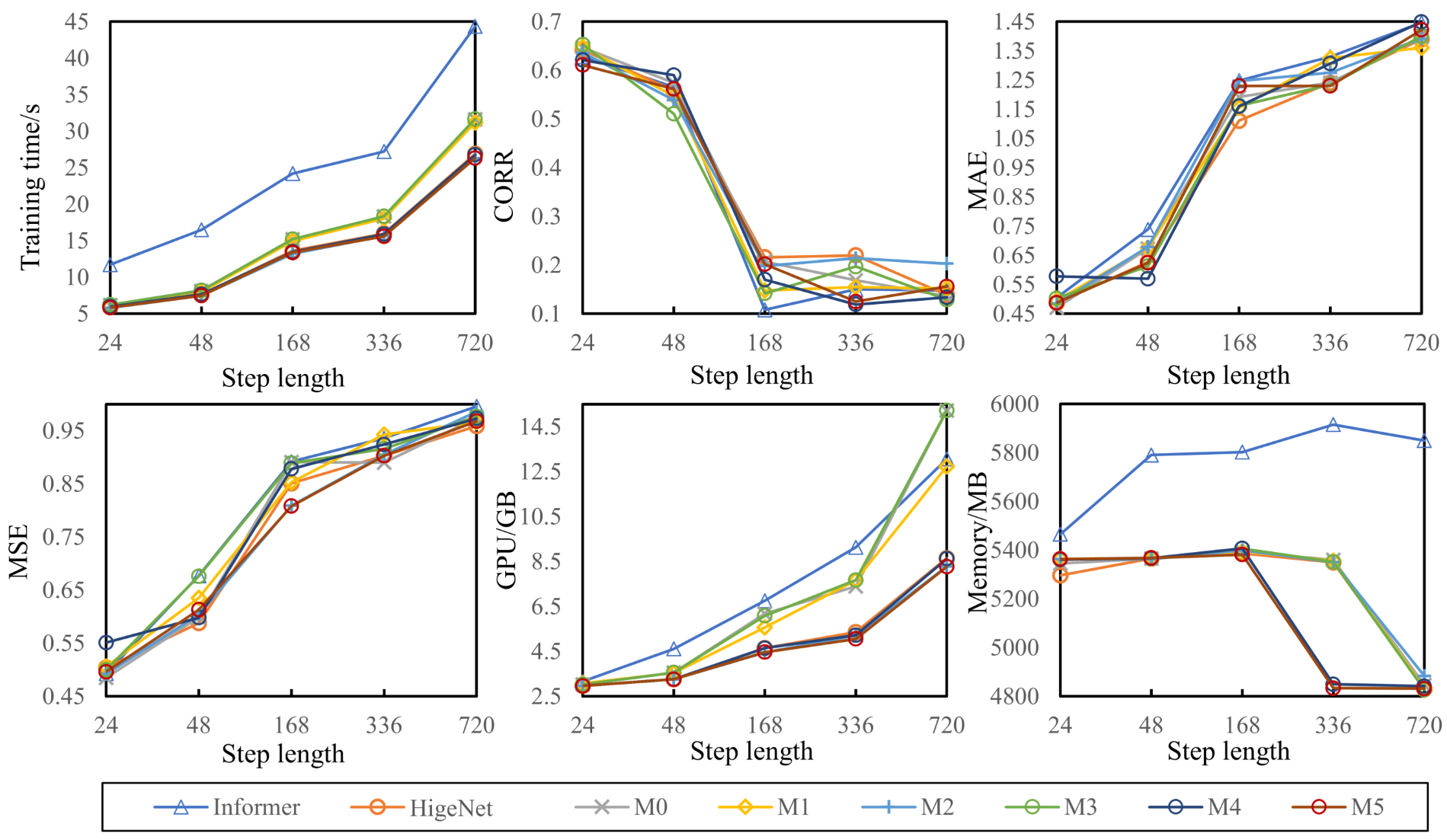}}
\caption{Multivariate long sequence time-series forecasting results for the ablation on ETTh$_1$ dataset}
\label{ETTh1_mtiablations}
\end{figure*}

In this case, we use all features for the experiments, including 7-dims for the ETT dataset and 20-dims for the AIOps dataset. We do not use the ARIMA algorithm due to its unavailability in the prediction of MTS. From Table \ref{multi}, we can observe that HigeNet has a significant advantage in the MTS prediction problem (count=40>11>3>0). Especially, for the ETTm$_1$ and AIOps datasets, our model has considerable strengths in all metrics. Among the other four methods Informer performs better and Wavenet+STM outperforms LSTNet and TPA-LSTM in general.

For such experimental results, we try to explain as follows : (1) Our embedding approach is more comprehensive, which enables the input representation to obtain more semantics in the MTS, such as the relationships between feature dimensions and the long- and short-term relationships,  and global temporal and positional couplings; (2) Our proposed distillation approach enables the single-branch network to retain more data characteristics and avoid the redundancy caused by module stacking; (3) The filtering of queries based on the learnable neural network ensures the continuity in the probability space.

\subsection{The performance comparison of different self-attention modules}
In a self-attention based model architecture, most of the time and resources are spent on the computation of the attention block. To further investigate the use of time and resources by different self-attention modules, we conduct the following experiments: Given three tensor matrices ($\mathbf{Q}$, $\mathbf{K}$ and $\mathbf{V}$), $\mathcal{A}(\mathbf{Q}, \mathbf{K}, \mathbf{V})$ is computed by three different self-attention methods, including NeuralSparse, ProbSparse from Informer and Canonical self-attention from Transormer. The time cost, GPU  and Memory usage are compared after the different methods are completed. We define the shape of $\mathbf{Q}$, $\mathbf{K}$ and $\mathbf{V}$ as (batch, seq\_len, heads, dims). The "batch" ranges form 1 to 64, and the "seq\_len" is between 1 and 1000. Additionally, the "heads" and "dims" are 8 and 64, respectively.

From Fig. \ref{self-attentionComps}, we can conclude as followings: (1) In general, the computation time used by various modules increases with the increase of the Seq\_len and batch. And as the batch rises, the ProbSparse method spends huge computation time, while the NeuralSparse and Canonical self-attention spend similar time. When batch $\ge$ 16 (at 16, 32 and 64), our NeuralSparse performs best with the increase of Seq\_len (over 800, 650 and 620). (2) Our method far outperforms the other two methods in GPU usage. (3) In terms of memory usage, all three models are similar and stable at around 4800MB.

\subsection{Ablation Study}
To compare the effect of different modules to our HigeNet, we design the ablation experiment in ETTh$_1$ dataset and the results are shown in Fig. \ref{ETTh1_uniablations} and Fig. \ref{ETTh1_mtiablations}. All tests and experimental environments are the same as described in the "Experimental Settings" section, except for $CORR$, $MAE$ and $MSE$, where we add training time, GPU and Memory usage. The horizontal coordinate of each of these sub-images is the step length of forecasting. 
In Fig. \ref{ETTh1_uniablations} and Fig. \ref{ETTh1_uniablations}, M0-M5 are the models after adding our proposed different modules and are represented by different colors in the figure, and they denote using embedding only, distillation only, NeuralSparse only, both embedding and distillation, both embedding and NeuralSparse, and both distillation and NeuralSparse, respectively. 

Through the ablation experiments, in terms of univariate time series prediction, we can deserve that : 
(1) The training time grows with the increase of prediction step length, among which Informer takes the longest time, almost more than twice as long as other models. 
(2) In regards to the $CORR$, all types of models have a declining trend of $CORR$ as the prediction step rises when it is less than 148. However, when the prediction step is greater 168, the $CORR$ starts to go up. At the same time, the prediction ability of Informer gradually diminishes, instead our HigeNet and other ablation models perform better.
(3) on $MAE$ and $MSE$, our HigeNet performs worse when the step length is 168 and 336, but at other step sizes, both our HigeNet and ablation models perform better than Informer's. In particular, when the step size is 720, our HigeNet is about 30\% more accurate than Informer.
(4) As for GPU usage, all models increase GPU usage as the prediction step length increases. Then, both Informer and the ablation model with ProbSparse self-attention module use higher GPU resources, however, our HigeNet and the ablation model with NeuralSparse self-attention module can save over 50\% of GPU computational resources compared to each other.
(5) Refer to memory usage, basically all the models use similar amounts, but the Informer model uses the most memory resources than the others.

In comparison with the univariate time series prediction, from the Fig. \ref{ETTh1_mtiablations}, we summarize the following: 
(1) There is a high degree of consistency across the measures, e.g., training time, $MAE$, $MSE$ and GPU usage all gradually increase as the prediction step length grows. But the memory usage shows a falling trend referring to our HigeNet and other ablation models. 
(2) Our model can still save a significant amount of training time and GPU resources, which are mainly attributed to our proposed embedding, distillation and NeuralSparse modules.
(3) The $CORR$, $MAE$ and $MSE$ change drastically in trend. In MTS, the accumulation of errors gradually increases as the prediction step length increases due to the increased number of data features and the complex relationship between individual features. However, in this case, our HigeNet is still optimal.

The above experiments prove that our proposed module has a positive contribution to the overall HigeNet performance and reaches the-state of-the-art.
\section{Conclusion and Discussion}
In this paper, we studied the multivariate long sequence
time series forecasting and proposed a corresponding deep learning method called HigeNet, which includes a more advanced embedding approach, a more concise distillation approach, and a self-attention mechanism that is more efficient in terms of computational time and computational resources. Extensive experiments were conducted on public dataset and real-world data we collected, and the experimental results demonstrated the effectiveness, efficiency and accuracy of our HigeNet in solving multivariate time series prediction problems.

\newpage
\appendix
\section{SUPPLEMENTARY MATERIAL}
\subsection{AIOps dataset introduction}
Our data comes from the operational logs of our actual online trading platform (All the data and experiment codes are available online\footnote{https://github.com/Torchlight-ljj/AIOPSdataset} and \footnote{https://github.com/Torchlight-ljj/HigeNet}), in addition to what is described in Table \ref{our_dataset} in the main body, a more detailed description is shown in the following Table \ref{addDescr}. The cols from 1 to 20 represent SP1A-DASD-RESP, SP1A-DASD-RATE, SP1B-DASD-RESP, SP1B-DASD-RESP, SP1C-DASD-RESP, SP1C-DASD-RATE, SP1D-DASD-RESP, SP1D-DASD-RATE, SP1A-MEM, SP1B-MEM, SP1C-MEM, SP1D-MEM, N-TASKS, TPS, SP1A-THOUT, SP1B-THOUT, SP1C-THOUT, SP1D-THOUT, SYSPLEX-MIPS and RESP-TIME, respectively.
\subsection{Experimental comparison of different data pre-processing methods}
Not only did we do the ablation experiments on the model, we also examined the effect of different data processing methods with the same experimental conditions as in the main body. We trained our HigeNet on the AIOps dataset and set the input length 96, the prediction step length 1 and the label length 48 (details in section "METHODOLOGY"). Additionally, the number of iterations is 8. The experimental results show that our model still has optimal performance, as shown in Table \ref{dataProcess}. We found it is optimal to standardize the training and test set by a dimension, so the ablation experiments were done based on this data processing method. Using original data to train is not feasible due to the large difference in the values of the features in our dataset.

\subsection{Ablation study in AIOps dataset}
To compare the effect of different modules to our HigeNet, we also designed the ablation experiment in our AIOps dataset and the results are shown in Table \ref{UniAblationAIOPS} and Table \ref{MulAblationAIOPS}. From the results, we can conclude that our HigeNet has a great advantage in accuracy and metrics, both in terms of univariate and multivariate time series prediction. As to training time and resource usage, both our model and the ablation models outperform the Informer model. In particular, the M2 model saves more training time and GPU usage due to using only our proposed NeuralSparse self-attention module, but does not perform Informer in terms of accuracy.

\subsection{Comparison of prediction effects of different methods}
\begin{figure}[t]
\centerline{\includegraphics[width=\linewidth]{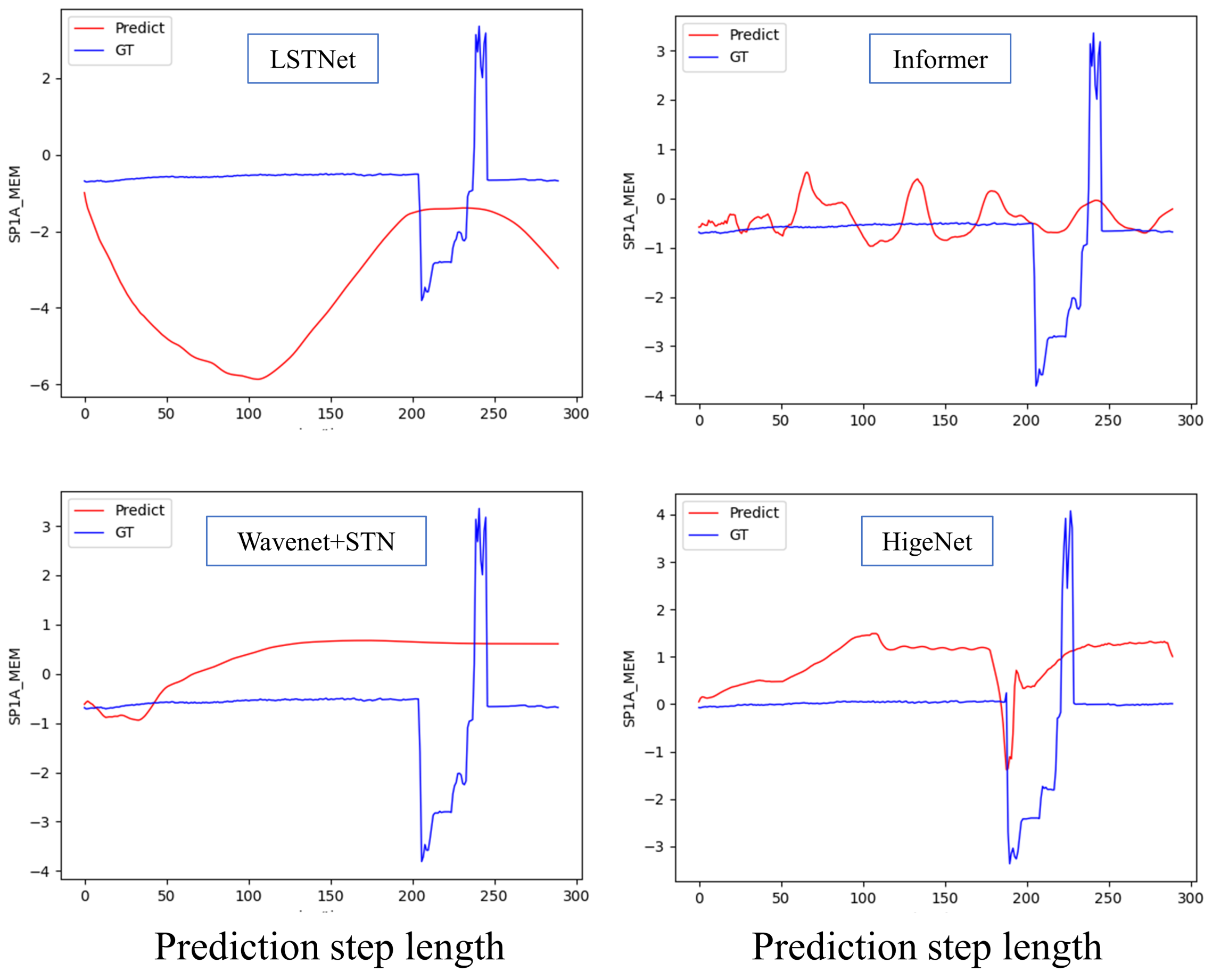}}
\caption{The performance comparison of different models.}
\label{modelsComps}
\end{figure}

To visualize the performance of our models, we randomly selected 288 consecutive data in the test set of AIOps dataset as input and predicted 288 steps in the future. The outputs of these four models (Wavenet+STN, LSTNet, Informer and HigeNet) are shown in the Fig. \ref{modelsComps}, where the red line indicates the ground truth and the blue line indicates the predicted values. In addition, the vertical coordinate of these 2 rows of sub-figures is SP1A-MEM.
\begin{table}[t]
\centering
\caption{The name and respective attributes of each column in the dataset, where Max is the maximum value, Min is the minimum value, M and V are the dataset's mean and variance, respectively. SP1A, SP1B, SP1C and SP1D are different logic partitions. RT, DRT, DT, NT, MU mean "Response time", "Disk response time", "Disk throughput", "Network throughput" and "Memory usage", respectively.
 }
 \label{addDescr}
% \centering
\setlength{\tabcolsep}{0.3mm}
\begin{tabular}{cccccc}
\hline
Cols & Max      & Min      & M       & V       & Description                \\ \hline
1       & 22.1     & 0.2      & 0.9203  & 0.67    & DRT of SP1A \\
2       & 29093.0  & 29.7     & 1755.0 & 1979.3 & DT of SP1A    \\
3       & 19.9     & 0.2      & 0.91    & 0.66    & DRT of SP1B \\
4       & 30160.0  & 28.0     & 1620.9 & 2021.9 & DT of SP1B    \\
5       & 21.6     & 0.2      & 0.90    & 0.63    & DRT of SP1C \\
6       & 28026.0  & 28.3     & 1651.1 & 2070.8 & DT of SP1C    \\
7       & 19.4     & 0.2      & 0.90    & 0.64    & DRT of SP1D \\
8       & 28661.0  & 29.3     & 1715.2 & 2213.6 & DT of SP1D    \\
9       & 97.84    & 3.93     & 41.94   & 4.23    & MU of SP1A       \\
10      & 82.46    & 4.54     & 40.76   & 3.72    & MU of SP1B       \\
11      & 84.08    & 3.98     & 40.40   & 3.64    & MU of SP1C       \\
12      & 89.64    & 3.98     & 40.50   & 3.61    & MU of SP1D       \\
13      & 50623.0  & 1.0      & 8112.7 & 7547.1 & Number of tasks            \\
14      & 843.71 & 0.0167   & 135.2  & 125.7  & Number of tasks/s \\
15      & 32.22    & 0.0      & 1.98    & 2.31    & NT of SP1A \\
16      & 24.99    & 0.0      & 1.54    & 1.97    & NT of SP1B \\
17      & 34.01    & 0.0      & 1.53    & 1.94    & NT of SP1C \\
18      & 24.28    & 0.0      & 1.53    & 1.94    & NT of SP1D \\
19      & 7596.0  & 102.03 & 1981.3 & 1692.5 & CPU load                   \\
20      & 0.5      & 0.0106   & 0.03    & 0.02    & RT of task      \\ \hline
\end{tabular}
\end{table}
\begin{table*}[t]
\centering
\caption{Ablation results of different data processing methods}
\label{dataProcess}
\begin{tabular}{c|cccc|c|c}
\hline
        & \multicolumn{4}{c|}{\begin{tabular}[c]{@{}c@{}}Unified processing of \\ test set and training set\end{tabular}}                                                                                                                                                                                                                                & \begin{tabular}[c]{@{}c@{}}Separate processing of \\ test set and training set\end{tabular} & No operation \\ \hline
Metrics & \multicolumn{1}{c|}{\begin{tabular}[c]{@{}c@{}}Normalized\\ by a dimension\end{tabular}} & \multicolumn{1}{c|}{\begin{tabular}[c]{@{}c@{}}Global\\ normalization\end{tabular}} & \multicolumn{1}{c|}{\begin{tabular}[c]{@{}c@{}}Global\\ standardization\end{tabular}} & \begin{tabular}[c]{@{}c@{}}Standardized by\\ a dimension\end{tabular} & \begin{tabular}[c]{@{}c@{}}Standardized \\ by a dimension\end{tabular}                      & --                  \\ \hline
CORR    & \multicolumn{1}{c|}{0.54}                                                                & \multicolumn{1}{c|}{0.11}                                                           & \multicolumn{1}{c|}{0.27}                                                             & \textbf{0.73}                                                         & 0.71                                                                                        & --                  \\ \hline
MAE     & \multicolumn{1}{c|}{0.03}                                                                & \multicolumn{1}{c|}{0.01}                                                           & \multicolumn{1}{c|}{0.12}                                                             & \textbf{0.3}                                                          & 0.33                                                                                        & --                  \\ \hline
MSE     & \multicolumn{1}{c|}{0.004}                                                               & \multicolumn{1}{c|}{0.001}                                                          & \multicolumn{1}{c|}{0.13}                                                             & \textbf{0.46}                                                         & 0.48                                                                                        & --                  \\ \hline
\end{tabular}
\end{table*}
\begin{table*}[t]
\centering
\caption{Univariate long sequence time-series forecasting results for the ablation on AIOps dataset}
\label{UniAblationAIOPS}
\begin{tabular}{c|ccc|ccc|ccc|c}
\hline
Metrics     & \multicolumn{3}{c|}{Training time (s)}               & \multicolumn{3}{c|}{GPU (GB)}                   & \multicolumn{3}{c|}{Memory (MB)}                          & \multirow{2}{*}{Count} \\ \cline{1-10}
Step length & 96             & 288              & 576              & 96             & 288            & 576           & 96                & 288               & 576               &                        \\ \hline
Informer    & 113.708        & 226.407          & 448.557          & 4.82           & 9.297          & 17.482        & 7297.824          & 5840.243          & 5854.266          & 0                      \\
HigeNet  & 68.887         & 128.261          & 321.83           & 3.51           & 5.928          & 9.848         & 5650.785          & 5643.758          & 5633.727          & 0                      \\
M0          & 73.204         & 147.131          & 373.914          & 3.922          & 7.732          & 15.043        & 5643.449          & \textbf{5615.895} & \textbf{5633.305} & 2                      \\
M1          & 71.815         & 145.768          & 371.337          & 3.996          & 7.578          & 13.725        & 5646.57           & 5617.152          & 5635.598          & 0                      \\
M2          & \textbf{66.79} & \textbf{124.926} & \textbf{316.551} & \textbf{3.41}  & \textbf{5.582} & \textbf{9.48} & 5645.055          & 5644.074          & 5647.383          & 6                      \\
M3          & 73.227         & 148.454          & 374.73           & 3.922          & 7.736          & 13.678        & 5646.352          & 5635.836          & 5637.738          & 0                      \\
M4          & 68.718         & 127.434          & 320.075          & 3.51           & 5.793          & 9.789         & \textbf{5642.445} & 5631.465          & 5634.926          & 1                      \\
M5          & 67.287         & 125.791          & 318.449          & 3.41           & 5.732          & 9.664         & 5647.238          & 5647.035          & 5643.23           & 0                      \\ \hline
Metrics     & \multicolumn{3}{c|}{CORR}                            & \multicolumn{3}{c|}{MSE}                        & \multicolumn{3}{c|}{MAE}                                  & \multirow{2}{*}{Count} \\ \cline{1-10}
Step length & 96             & 288              & 576              & 96             & 288            & 576           & 96                & 288               & 576               &                        \\ \hline
Informer    & 0.24           & 0.266            & 0.25             & 1.323          & 1.329          & 1.35          & 0.321             & 0.293             & 0.307             & 0                      \\
HigeNet  & \textbf{0.271} & \textbf{0.306}   & \textbf{0.287}   & \textbf{1.296} & 1.3            & \textbf{1.30} & \textbf{0.309}    & 0.283             & \textbf{0.278}    & 7                      \\
M0          & 0.241          & 0.287            & 0.27             & 1.318          & 1.304          & 1.33          & 0.333             & 0.295             & 0.303             & 0                      \\
M1          & 0.237          & 0.304            & 0.266            & 1.326          & \textbf{1.286} & 1.333         & 0.327             & \textbf{0.276}    & 0.301             & 2                      \\
M2          & 0.269          & 0.274            & 0.283            & 1.312          & 1.307          & 1.302         & 0.313             & 0.291             & 0.287             & 0                      \\
M3          & 0.262          & 0.298            & 0.269            & 1.31           & 1.3            & 1.356         & 0.313             & 0.283             & 0.303             & 0                      \\
M4          & 0.259          & 0.294            & 0.272            & 1.317          & 1.295          & 1.334         & 0.319             & 0.284             & 0.291             & 0                      \\
M5          & 0.23           & 0.289            & 0.266            & 1.324          & 1.329          & 1.329         & 0.336             & 0.303             & 0.3               & 0                      \\ \hline
\end{tabular}
\end{table*}
\begin{table*}[t]
\centering
\caption{Multivariate long sequence time-series forecasting results for the ablation on AIOps dataset}
\label{MulAblationAIOPS}
\begin{tabular}{c|ccc|ccc|ccc|c}
\hline
Metrics     & \multicolumn{3}{c|}{Training time (s)}                & \multicolumn{3}{c|}{GPU (GB)}                    & \multicolumn{3}{c|}{Memory (MB)}                         & \multirow{2}{*}{Count} \\ \cline{1-10}
Step length & 96              & 288              & 576              & 96             & 288            & 576            & 96                & 288               & 576              &                        \\ \hline
Informer    & 114.07          & 295.135          & 451.903          & 4.23           & 10.518         & 16.166         & 5776.312          & 5934.793          & 5941.969         & 0                      \\
HigeNet  & 70.169          & 196.917          & 324.762          & 3.457          & 7.109          & 9.848          & 5765.066          & 6665.047          & 5707.625         & 0                      \\
M0          & 74.059          & 227.702          & 377.622          & 3.93           & 9.148          & 15.041         & \textbf{5691.383} & 6683.336          & 7457.691         & 1                      \\
M1          & 72.997          & 227.302          & 374.595          & 3.795          & 9.113          & 14.391         & 5990.918          & \textbf{5684.297} & 5753.844         & 1                      \\
M2          & \textbf{68.41} & \textbf{193.097} & \textbf{320.86} & \textbf{3.406} & \textbf{6.969} & \textbf{9.609} & 5713.188          & 5707.047          & 5757.539         & 6                      \\
M3          & 74.738          & 229.749          & 378.727          & 3.893          & 9.514          & 13.971         & 5698.602          & 6177.074          & 5749.953         & 0                      \\
M4          & 70.026          & 195.613          & 323.241          & 3.461          & 7.166          & 9.947          & 5711.855          & 5693.969          & \textbf{5693.66} & 1                      \\
M5          & 68.414          & 194.728          & 320.864          & 3.414          & 7.025          & 9.617          & 5730.762          & 5709.98           & 6974.969         & 0                      \\ \hline
Metrics     & \multicolumn{3}{c|}{CORR}                             & \multicolumn{3}{c|}{MSE}                         & \multicolumn{3}{c|}{MAE}                                 & \multirow{2}{*}{Count} \\ \cline{1-10}
Step length & 96              & 288              & 576              & 96             & 288            & 576            & 96                & 288               & 576              &                        \\ \hline
Informer    & 0.572           & 0.558            & 0.506            & 0.683          & 0.719          & 0.775          & 0.404             & 0.403             & 0.434            & 0                      \\
HigeNet  & \textbf{0.627}  & \textbf{0.588}   & \textbf{0.542}   & 0.64           & \textbf{0.653} & \textbf{0.711} & 0.411             & \textbf{0.373}    & 0.432            & 6                      \\
M0          & 0.605           & 0.497            & 0.53             & 0.638          & 0.824          & 0.741          & 0.402             & 0.428             & \textbf{0.41}    & 1                      \\
M1          & 0.582           & 0.513            & 0.545            & 0.662          & 0.758          & 0.731          & 0.422             & 0.42              & 0.431            & 0                      \\
M2          & 0.591           & 0.568            & 0.537            & 0.652          & 0.685          & 0.734          & 0.404             & 0.398             & 0.429            & 0                      \\
M3          & 0.606           & 0.554            & 0.538            & 0.631          & 0.696          & 0.717          & 0.401             & 0.407             & 0.431            & 0                      \\
M4          & 0.599           & 0.533            & 0.555            & 0.652          & 0.745          & 0.692          & 0.408             & 0.428             & 0.411            & 0                      \\
M5          & 0.618           & 0.55             & 0.508            & \textbf{0.624} & 0.714          & 0.771          & \textbf{0.383}    & 0.398             & 0.448            & 2                      \\ \hline
\end{tabular}
\end{table*}

% \balance
\end{document}